\documentclass{fairmeta}
\usepackage{amsmath}
\usepackage{enumerate} 
\usepackage{bm}
\usepackage{algorithm}
\usepackage{floatflt}
\usepackage{algpseudocode}
\usepackage{amsfonts}
\usepackage{amsthm}
\usepackage{newtxtt}
\usepackage{algorithm}
\usepackage{colortbl} 
\usepackage{cleveref}
\usepackage{diagbox} 
\usepackage[utf8]{inputenc}
\usepackage{textgreek}
\usepackage{colortbl}
\usepackage{nicematrix}
\usepackage{makecell}
\usepackage{float}
\usepackage{arydshln}
\usepackage[frozencache,cachedir=.]{minted}
\usepackage{caption}
\usepackage{subcaption}
\usepackage{tcolorbox}
\usepackage{amssymb}
\usepackage{xspace}
\usepackage{wrapfig}
\usepackage{adjustbox}
\usepackage{tabularx}
\usepackage{booktabs}
\usepackage{mathtools}
\usepackage{wrapfig}
\usepackage{amssymb}
\usepackage{graphicx}
\usepackage{makecell}
\usepackage{booktabs,tabularx,array}

\usepackage{silence}
\makeatletter
\patchcmd{\wrong@fontshape}{\@gobbletwo}{}{}{}
\makeatother
\definecolor{upColor}{RGB}{17,138,21}
\definecolor{downColor}{RGB}{174,36,67}

\newtheorem{theorem}{Theorem}[]

\newtheorem{remark1}[theorem]{Remark}

\title{Single-Pixel Vision-Language Model for Intrinsic Privacy-Preserving Behavioral Intelligence}
\author[]{Hongjun~An$^\dagger$}
\author[]{Yiliang~Song$^\dagger$}
\author[]{Jiawei~Shao}
\author[]{Zhe~Sun}
\author[]{Xuelong~Li$^*$}
\affiliation[]{Institute of Artificial Intelligence (TeleAI), China Telecom}
\contribution{\textsuperscript{$\dagger$}Equal Contributions$\quad$\textsuperscript{*}Corresponding Author}
\metadata[Keywords]{Single-pixel sensing, privacy preservation, behavioral intelligence, ethical monitoring}

\metadata[Correspondence to]{Xuelong Li (\email{xuelong\_li@ieee.org})}

\begin{document}

\abstract{
Adverse social interactions, such as bullying, harassment, and other illicit activities, pose significant threats to individual well-being and public safety, leaving profound impacts on physical and mental health.
However, these critical events frequently occur in privacy-sensitive environments like restrooms, and changing rooms, where conventional surveillance is prohibited or severely restricted by stringent privacy regulations and ethical concerns.
Here, we propose the Single-Pixel Vision-Language Model (SP-VLM), a novel framework that reimagines secure environmental monitoring. 
It achieves intrinsic privacy-by-design by capturing human dynamics through inherently low-dimensional single-pixel modalities and inferring complex behavioral patterns via seamless vision-language integration.
Building on this framework, we demonstrate that single-pixel sensing intrinsically suppresses identity recoverability, rendering state-of-the-art face recognition systems ineffective below a critical sampling rate.
We further show that SP-VLM can nonetheless extract meaningful behavioral semantics, enabling robust anomaly detection, people counting, and activity understanding from severely degraded single-pixel observations.
Combining these findings, we identify a practical sampling-rate regime in which behavioral intelligence emerges while personal identity remains strongly protected.
Together, these results point to a human-rights-aligned pathway for safety monitoring that can support timely intervention without normalizing intrusive surveillance in privacy-sensitive spaces.
}

\maketitle

\section{Introduction}
% 第一段：社会背景与痛点 - 修正版 v2
The severe challenges posed by adverse social interactions, ranging from bullying and harassment to other illicit activities, represent a profound and growing threat to individual well-being and public safety globally~\cite{WHO2024YouthViolence}. 
These incidents, leaving indelible marks on both physical and mental health, underscore an urgent societal need for effective protective and intervention strategies. 
However, a critical dilemma emerges in environments where these vulnerabilities are often highest: privacy-sensitive spaces such as public restrooms, changing facilities, and secluded corridors. 
In these confined and private settings, conventional surveillance methods are explicitly prohibited or severely restricted by stringent privacy regulations and deep-seated ethical concerns, rendering victims vulnerable and preventing timely detection or intervention. 
% This creates a significant gap in our ability to foster safer communities while upholding fundamental human rights, leaving critical incidents harder to detect and respond to in a timely manner.
This creates a critical challenge in fostering safer communities while balancing the protection of fundamental human rights, making timely detection and intervention in privacy-sensitive environments increasingly difficult.

\begin{figure}[ht]
    \centering
    \includegraphics[width=\textwidth]{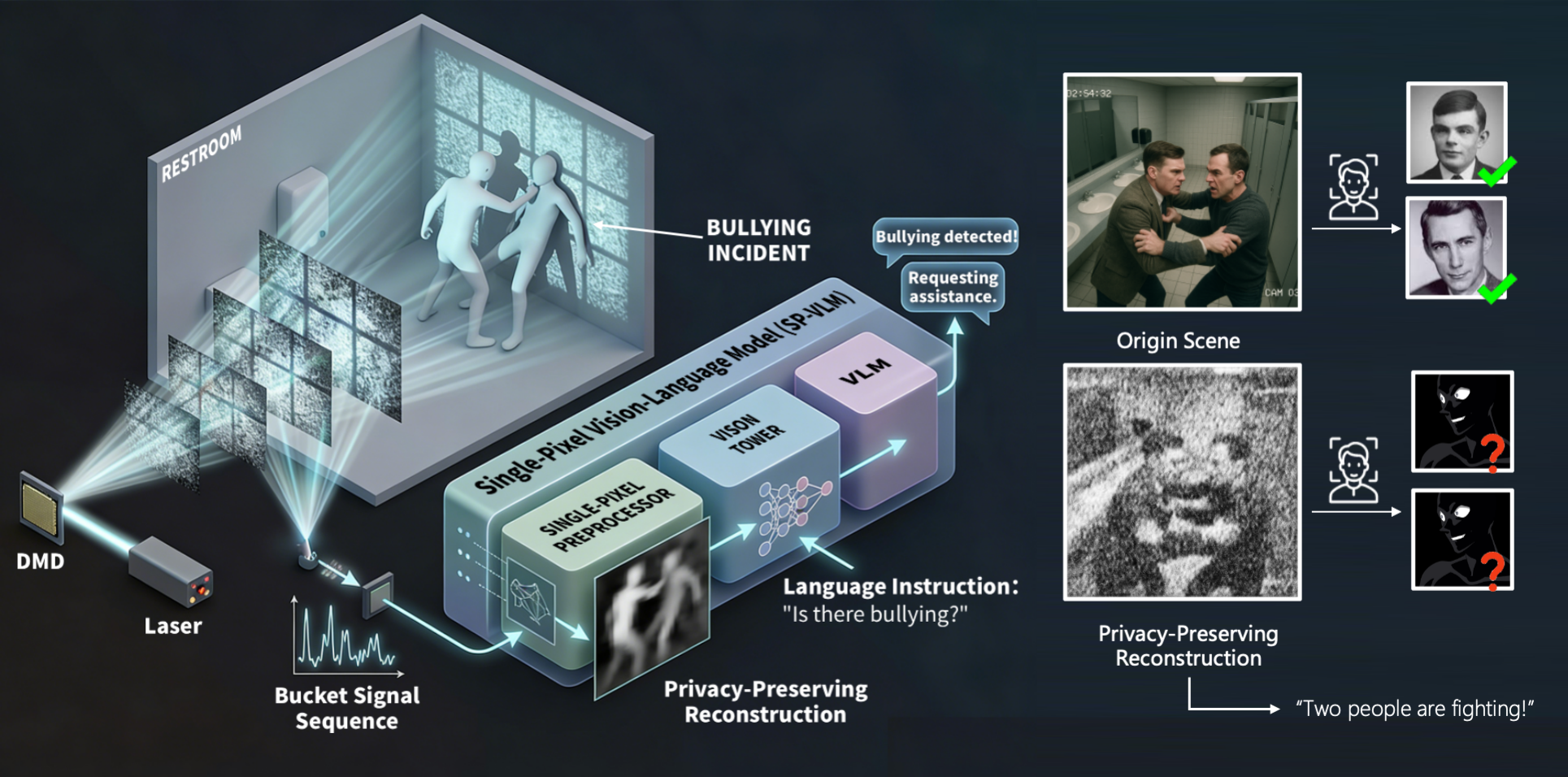}
    \caption{\textbf{Overview of the Single-Pixel Vision-Language Model (SP-VLM) framework.} The system utilizes active single-pixel imaging to capture optical bucket signals from privacy-sensitive spaces, avoiding the acquisition of high-fidelity images. The SP-VLM processes these low-dimensional signals to generate privacy-preserving reconstructions that mask facial identities while retaining behavioral dynamics. By integrating visual data with language instructions, the model achieves robust semantic reasoning to detect anomalies without compromising individual privacy.}
    \label{fig:main}
\end{figure}

Addressing this challenge, two distinct technological frontiers offer tantalizing yet incomplete solutions. 
% On one hand, single-pixel imaging (SPI) has emerged as a disruptive computational sensing paradigm, capable of reconstructing visual scenes from sparse, indirect light measurements~\cite{duarte2008single, li2024part}. 
On one hand, single-pixel imaging (SPI) represents an innovative computational sensing paradigm that reconstructs visual scenes from sparse, indirect light measurements~\cite{duarte2008single, li2024part}.
Its fundamental operation, involving modulating light and detecting total reflected intensity with a single photodetector, uniquely enables image acquisition even under inherently low-dimensional sensing modalities~\cite{chen2023computational, chen2025large}. 
This characteristic inherently limits the reconstruction of high-fidelity details, making SPI a potential candidate for privacy-preserving applications. 
On the other hand, recent advancements in artificial intelligence have led to sophisticated vision-language models (VLMs), which excel at integrating visual and textual information to achieve deep semantic understanding and reasoning about complex scenes and human activities, revolutionizing various applications from image captioning to advanced behavioral recognition~\cite{OpenAI2025GPT5, TeamGemini2023, an2025ai}. 
% A critical synergy has yet to be realized: while SPI naturally supports privacy-preserving sensing, ultra-low-rate measurements are dominated by noise and structural loss, and existing approaches do not provide sufficiently robust models for interpreting human behaviors from such degraded signals.
A critical gap remains in current methods: while SPI inherently supports privacy-preserving sensing, ultra-low-rate measurements introduce significant noise and loss of structural integrity, which existing models struggle to overcome in effectively interpreting human behaviors from such degraded signals.
While existing high-performance VLMs exhibit remarkable capabilities in semantic reasoning, they generally depend on high-fidelity visual inputs, which presents ethical challenges for their application in privacy-sensitive settings where detailed visual information is intentionally avoided or constrained.
% Meanwhile, existing high-performance VLMs, despite their unparalleled semantic capabilities, fundamentally rely on rich, high-fidelity visual inputs, making their direct deployment ethically problematic in privacy-sensitive environments where detailed visual information is intentionally withheld.

% 第三段 (保持不变)
% Here, we propose the Single-Pixel Vision-Language Model (SP-VLM), a paradigm-shifting approach designed to resolve this fundamental tension between privacy and perceptive intelligence. 
We propose the Single-Pixel Vision-Language Model (SP-VLM), a novel framework that addresses the critical tension between privacy preservation and perceptual intelligence.
Our framework redefines secure environmental monitoring by seamlessly integrating the intrinsic privacy-preserving capabilities of single-pixel sensing with the advanced semantic reasoning of tailored vision-language architectures. 
SP-VLM achieves this by precisely capturing human dynamics through inherently low-dimensional single-pixel modalities, which intrinsically safeguard identity information, and then inferring complex behavioral patterns via sophisticated, privacy-aware vision-language integration (Fig.~\ref{fig:main}). 
This novel methodology thus delivers intrinsic privacy-preserving behavioral intelligence, enabling ethically grounded anomaly detection in previously unmonitorable spaces and positioning low-dimensional sensing as a practical foundation for resource-efficient, intrinsically secure AI perception.

\section{Results}
% 总括性引言段落 (保持不变)
In this study, we conducted a comprehensive evaluation of the proposed SP-VLM, aiming to quantitatively validate its capability to achieve advanced behavioral intelligence while maintaining intrinsic privacy preservation. Through a series of simulated experiments, we first demonstrated the unidentifiability of individual identities under low single-pixel sampling rates, and subsequently revealed its effectiveness in capturing and decoding human behavioral patterns across varying sampling rates. Ultimately, we delineated a unique sampling rate regime where the model reliably infers complex behaviors while strictly safeguarding personal identity.

\subsection{Single-Pixel Sensing Provides Intrinsic Identity Protection}

To rigorously verify the identity unidentifiability guaranteed by single-pixel (SP) sensing, we conducted controlled face recognition experiments using a state-of-the-art commercial-grade face recognition system, whose advertised verification accuracy reaches 99.99\%, substantially exceeding human-level face recognition performance (97.52\%)~\cite{LuTang2014GaussianFace}. Experiments were performed on the Labeled Faces in the Wild (LFW) dataset~\cite{Huang2007LFW}. All face images were uniformly resized to a spatial resolution of $320\times320$ pixels prior to evaluation. We selected individuals with at least two face images, enrolling one image per identity into the face gallery and using a different image from the same individual for probe-based identification.

When evaluated on original RGB images, the recognition system achieved 100.00\% Top-1 identification accuracy, confirming the reliability and strength of the recognition model. We then applied the single-pixel preprocess module of SP-VLM to reconstruct face images at varying sensing resolutions, using 1000, 2000, 3000, and 4000 speckle patterns. The reconstructed outputs were subsequently fed into the same face recognition system under identical evaluation settings.

\begin{table}[h]
\centering
\caption{\textbf{Top-1 face identification accuracy under single-pixel reconstructions.}
We evaluate a commercial-grade face recognition system on LFW faces reconstructed from single-pixel measurements with different numbers of speckle patterns (columns).
Results are reported across varying gallery sizes (rows). The last row reports the mean and standard deviation (std) computed across gallery sizes.}
\label{tab:identity_protection_top1}
\setlength{\tabcolsep}{8pt}
\renewcommand{\arraystretch}{1.15}
\begin{tabularx}{\textwidth}{l *{4}{>{\centering\arraybackslash}X}}
\toprule
\textbf{Gallery size} & \textbf{1000} & \textbf{2000} & \textbf{3000} & \textbf{4000} \\
\midrule
100 identities  & 0.0000 & 0.0000 & 0.0500 & 0.1700 \\
200 identities  & 0.0000 & 0.0000 & 0.0350 & 0.1250 \\
300 identities  & 0.0000 & 0.0000 & 0.0300 & 0.1170 \\
400 identities  & 0.0000 & 0.0000 & 0.0375 & 0.1275 \\
\midrule
\textbf{Mean $\pm$ std} &
\textbf{0.0000 $\pm$ 0.0000} &
\textbf{0.0000 $\pm$ 0.0000} &
\textbf{0.0381 $\pm$ 0.0085} &
\textbf{0.1349 $\pm$ 0.0238} \\
\bottomrule
\end{tabularx}
\vspace{-0.6em}
\end{table}

The results (Table~\ref{tab:identity_protection_top1}) reveal a clear and consistent positive correlation between recognition accuracy and the number of speckle patterns. Crucially, when the speckle count remains below 3000, identity recognition accuracy drops sharply to below 5\%, indicating a near-complete failure of identity inference. This trend was further validated through robustness analyses across different gallery sizes, with face databases containing 100, 200, 300, and 400 enrolled identities, respectively. In all configurations, the same conclusion holds: single-pixel sensing intrinsically suppresses identity-related information at low sampling rates.

To rule out potential bias induced by strict Top-1 evaluation, we additionally report Top-2 through Top-5 identification accuracies. The results remain quantitatively consistent, exhibiting no meaningful recovery of identity recognition under low speckle regimes. Collectively, these findings demonstrate that SP sensing provides a strong, architecture-agnostic form of identity protection, confirming its suitability as a foundational sensing mechanism for intrinsically privacy-preserving behavioral intelligence.

\subsection{SP-VLM Achieves Behavioral Intelligence from Single-Pixel Sensing}

\begin{figure}[ht]
    \centering
    \includegraphics[width=\textwidth]{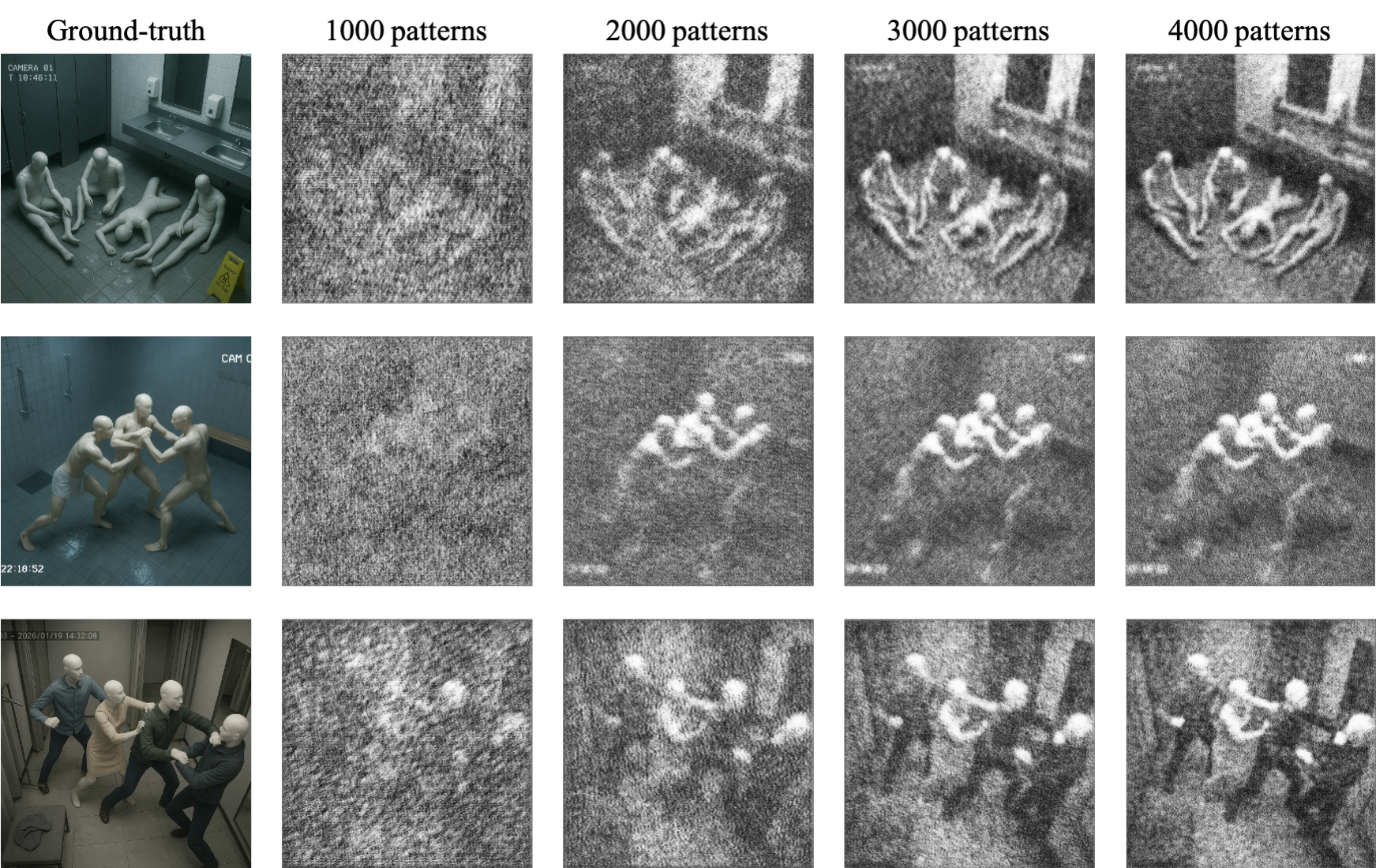}
    \caption{Representative examples of privacy-sensitive scenes and their corresponding single-pixel reconstructions at increasing sampling rates.}
    \label{fig:exp2}
\end{figure}

\begin{table}[ht]
\centering
\caption{\textbf{Behavioral intelligence on the synthetic privacy-sensitive dataset (test set).}
We compare Qwen3-VL-8B against SP-VLM (8B) under different numbers of single-pixel sampling patterns.}
\label{tab:spvlm_behavioral_intelligence}
\setlength{\tabcolsep}{5pt}
\renewcommand{\arraystretch}{1.15}
\begin{tabularx}{\textwidth}{l *{5}{>{\centering\arraybackslash}X}}
\toprule
\textbf{Metric} &
\textbf{Qwen3-VL-8B} &
\textbf{SP-VLM (1000)} &
\textbf{SP-VLM (2000)} &
\textbf{SP-VLM (3000)} &
\textbf{SP-VLM (4000)} \\
\midrule
Anomaly Detection (Accuracy) &
40.52\% & 41.38\% & 57.76\% & 72.41\% & 76.72\% \\
People Counting (MAE) &
0.7328 & 1.6379 & 0.9828 & 0.8534 & 0.6897 \\
Scene Recognition (ROUGE-L) &
0.0198 & 0.6099 & 0.6239 & 0.6187 & 0.6224 \\
Human Activity Recognition (ROUGE-L) &
0.0462 & 0.0314 & 0.2075 & 0.2067 & 0.2544 \\
\bottomrule
\end{tabularx}
\vspace{-0.6em}
\end{table}

Privacy regulations and ethical constraints make it difficult to collect large-scale behavior datasets in privacy-sensitive environments, limiting quantitative evaluation of behavioral intelligence in such spaces. Building on the previous subsection, where we show that SP sensing substantially suppresses identity recoverability below a critical sampling rate, we construct a synthetic privacy-sensitive activity dataset using an AI-based image generation pipeline. The dataset contains 2,000 images spanning public restrooms, changing rooms, and shared shower facilities (Fig.~\ref{fig:exp2}), covering both normal and abnormal behaviors (details and prompt templates are provided in the Appendix \ref{syn_dataset}). To avoid identity leakage, all generated humans use anonymized dummy faces (realistic identity-specific faces are typically rejected by generative systems in private scenarios). We split the data into training/validation/test sets with an 8:1:1 ratio.

We instantiate SP-VLM with an 8B backbone and compare it to a parameter-matched baseline, Qwen3-VL-8B~\cite{Qwen3-VL}, on the held-out test set. We evaluate four tasks: anomaly detection (Accuracy), people counting (MAE), scene recognition (ROUGE-L~\cite{lin-2004-rouge}), and human activity recognition (ROUGE-L). As summarized in Table~\ref{tab:spvlm_behavioral_intelligence}, performance improves consistently with the number of SP sampling patterns. Notably, once patterns exceed 2,000, anomaly detection accuracy rises above 50\%, indicating that behavior-level semantics are recoverable even in identity-obfuscating sensing regimes. At higher sampling rates, SP-VLM achieves strong performance across tasks (e.g., 76.72\% anomaly detection accuracy at 4,000 patterns) while maintaining low identity recoverability as shown previously, delineating a practical operating regime for intrinsically privacy-preserving behavioral intelligence.

\subsection{A Practical Sampling-Rate Interval Balances Identity Privacy and Behavioral Intelligence}

Taken together, the above two experiments reveal a non-trivial operating regime in which behavioral intelligence and identity privacy are simultaneously achievable. On the one hand, results from the identity protection study demonstrate that, under the same experimental settings, identity recognition accuracy remains near chance level when the number of single-pixel sampling patterns is below approximately 3,000 (Table \ref{tab:identity_protection_top1}). On the other hand, the behavioral intelligence evaluation shows that meaningful semantic understanding, particularly for anomaly detection and activity recognition, emerges once the sampling rate exceeds roughly 2,000 patterns (Table \ref{tab:spvlm_behavioral_intelligence}).

This overlap delineates a practical trade-off regime, empirically spanning approximately 2,000-3,000 single-pixel patterns in our setup, where SP-VLM is able to reliably infer behavior-level semantics while identity-related cues remain strongly suppressed. Within this interval, the model achieves anomaly detection accuracy above 50\% and steadily improving scene and activity recognition performance, without a corresponding recovery of identity recognizability as evidenced by face recognition failure across multiple gallery sizes.

Importantly, this regime does not represent a hard boundary but rather a task- and model-dependent operating window, shaped by sensing resolution, reconstruction quality, and downstream semantic complexity. Nevertheless, its existence highlights a key insight: privacy preservation and behavioral intelligence need not be mutually exclusive. Instead, low-dimensional sensing combined with end-to-end semantic modeling enables a controllable balance between information utility and privacy risk.

% \begin{figure}[h]
%     \centering
%     \includegraphics[width=0.8\textwidth]{figures/privacy_intelligence_balance.pdf}
%     \caption*{\textbf{Fig. 3 | Relationship between single-pixel sampling rate, identity security, and behavioral intelligence.}}
%     \label{fig:balance_regime}
% \end{figure}

\section{Discussion}

This study proposes the SP-VLM, a novel framework that balances privacy preservation and behavioral intelligence in privacy-sensitive environments. By combining SPI with VLMs, we propose a framework capable of capturing human dynamics through low-dimensional signals while preserving personal identity. Unlike traditional vision-based systems, which often rely on high-resolution images, SP-VLM enables robust behavioral analysis without compromising individual privacy. Our experiments demonstrate that, even under severely degraded sensing conditions, SP-VLM can reliably perform tasks such as anomaly detection, activity recognition, and people counting, establishing a new paradigm for ethical monitoring in restricted spaces.

While the use of synthetic datasets offers valuable insights, it also introduces certain limitations, primarily the lack of diversity and complexity found in real-world environments. As the next step, validating SP-VLM in uncontrolled, real-world settings will be crucial to fully assess its applicability and robustness across a wider range of human behaviors and environmental factors.

Looking forward, there are several avenues for extending the impact of SP-VLM. One key direction is the validation of SP-VLM in real-world, uncontrolled environments to test its practical applicability in everyday privacy-sensitive spaces. Future research should also explore expanding the model’s capacity to handle a broader variety of tasks, such as more complex human-object interactions or cross-modal event detection. Moreover, as AI-driven solutions continue to grow, SP-VLM’s ability to balance privacy with behavioral intelligence presents exciting opportunities in areas like smart city surveillance, healthcare monitoring, and personal security. These applications promise a future where privacy and safety can coexist seamlessly, providing security without infringing on individual rights.

\section{Methods}

\subsection{System and notation}

A scene is discretized on a spatial grid and vectorized as
\begin{equation}
\mathbf{x}\in\mathbb{R}^{N},\qquad N=H\cdot W.
\end{equation}
A library of illumination patterns is defined as
\begin{equation}
\mathcal{S}=\{\boldsymbol{\phi}_i\}_{i=1}^{n},\qquad \boldsymbol{\phi}_i\in\mathbb{R}^{N},
\end{equation}
and stacked into a pattern matrix
\begin{equation}
\mathbf{\Phi}=
\begin{bmatrix}
\boldsymbol{\phi}_1^\top\\
\vdots\\
\boldsymbol{\phi}_n^\top
\end{bmatrix}\in\mathbb{R}^{n\times N}.
\end{equation}
During an observation window we use $M$ patterns and define the sampling rate as
\begin{equation}
\rho \triangleq \frac{M}{N}\in(0,1].
\end{equation}
Pattern selection is represented by a binary selection matrix $\mathbf{S}\in\{0,1\}^{M\times n}$ with exactly one nonzero per row, yielding
\begin{equation}
\mathbf{\Phi}_M=\mathbf{S}\mathbf{\Phi}\in\mathbb{R}^{M\times N}.
\end{equation}
This matrix form isolates the effect of sampling rate: the pattern library is fixed while the sensing operator changes only through its selected rows.

\subsection{Optical measurement model}
For the $m$-th projected pattern, the bucket detector outputs a scalar integral of the returned intensity. The discrete forward model is
\begin{equation}
y_m=\langle \boldsymbol{\phi}_m,\mathbf{x}\rangle+\varepsilon_m
=\sum_{j=1}^{N}\phi_{m,j}x_j+\varepsilon_m,\qquad m=1,\dots,M,
\end{equation}
where $\varepsilon_m$ aggregates detector noise, residual ambient illumination, and slow system drift. Stacking all measurements gives
\begin{equation}
\mathbf{y}=
\begin{bmatrix}
y_1\\ \vdots\\ y_M
\end{bmatrix}
=\mathbf{\Phi}_M\mathbf{x}+\boldsymbol{\varepsilon}.
\end{equation}

To suppress direct-current bias and ambient drift, we use effective zero-mean patterns. One implementation uses complementary modulation. Let $\bar{\boldsymbol{\phi}}_m=\mathbf{1}-\boldsymbol{\phi}_m$ and record two bucket responses, then take their difference. This is equivalent to the model
\begin{equation}
y_m^{\Delta}=\langle \tilde{\boldsymbol{\phi}}_m,\mathbf{x}\rangle+\varepsilon_m,\qquad
\tilde{\boldsymbol{\phi}}_m\triangleq 2\boldsymbol{\phi}_m-\mathbf{1}.
\end{equation}
In what follows, $\mathbf{y}$ and $\mathbf{\Phi}_M$ denote the calibrated, zero-mean, energy-normalized effective measurements and sensing matrix.

\subsection{Multi-frame stacking and spatiotemporal operator}
Behaviour recognition depends on temporal dynamics. Consider an observation window with $T$ frames. Define
\begin{equation}
\mathbf{X}=[\mathbf{x}^{(1)},\dots,\mathbf{x}^{(T)}]\in\mathbb{R}^{N\times T},\qquad
\mathbf{Y}=[\mathbf{y}^{(1)},\dots,\mathbf{y}^{(T)}]\in\mathbb{R}^{M\times T}.
\end{equation}
If the same $\mathbf{\Phi}_M$ is used across frames, then
\begin{equation}
\mathbf{Y}=\mathbf{\Phi}_M\mathbf{X}+\mathbf{E}.
\end{equation}
Vectorizing yields a Kronecker structured operator
\begin{equation}
\mathrm{vec}(\mathbf{Y})=
(\mathbf{I}_T\otimes \mathbf{\Phi}_M)\,\mathrm{vec}(\mathbf{X})
+\mathrm{vec}(\mathbf{E}),
\end{equation}
where $\otimes$ denotes the Kronecker product. This representation supports both direct inference in the measurement domain and reconstruction with spatiotemporal priors.

\subsection{Calibration and noise whitening}
The acquisition chain is modeled as an affine transformation of the ideal measurement:
\begin{equation}
\mathbf{y}_{\mathrm{raw}}=\mathbf{G}\mathbf{y}+\mathbf{o},\qquad
\mathbf{G}=\mathrm{diag}(g_1,\dots,g_M).
\end{equation}
We estimate the offset $\mathbf{o}$ from dark measurements and the gain matrix $\mathbf{G}$ from reference measurements, then calibrate by
\begin{equation}
\mathbf{y}=\mathbf{G}^{-1}(\mathbf{y}_{\mathrm{raw}}-\mathbf{o}).
\end{equation}
We model noise as
\begin{equation}
\boldsymbol{\varepsilon}\sim \mathcal{N}(\mathbf{0},\mathbf{\Sigma}).
\end{equation}
To remove correlated noise and heteroscedasticity, we apply whitening
\begin{equation}
\tilde{\mathbf{y}}=\mathbf{\Sigma}^{-1/2}\mathbf{y},\qquad
\tilde{\mathbf{\Phi}}_M=\mathbf{\Sigma}^{-1/2}\mathbf{\Phi}_M,
\end{equation}
which yields an equivalent white-noise model
\begin{equation}
\tilde{\mathbf{y}}=\tilde{\mathbf{\Phi}}_M\mathbf{x}+\tilde{\boldsymbol{\varepsilon}},\qquad
\tilde{\boldsymbol{\varepsilon}}\sim \mathcal{N}(\mathbf{0},\mathbf{I}_M).
\end{equation}
For notational simplicity, we continue to write $\mathbf{y}$ and $\mathbf{\Phi}_M$ for the standardized quantities.

\subsection{Inference tasks and conditional probability formulation}
We consider two recognisability objectives. Privacy recognisability targets identity-related facial information. Behaviour recognisability targets risk-related actions. Let $U\in\mathcal{U}$ denote a privacy label and $B\in\mathcal{B}$ denote a behaviour label. We define measurement-domain posteriors
\begin{equation}
p_{\theta_{\mathrm{priv}}}(u\mid \mathbf{y},\rho),\qquad
p_{\theta_{\mathrm{beh}}}(b\mid \mathbf{Y},\rho),
\end{equation}
and use maximum a posteriori decisions
\begin{equation}
\hat{U}=\arg\max_{u\in\mathcal{U}}p_{\theta_{\mathrm{priv}}}(u\mid \mathbf{y},\rho),\qquad
\hat{B}=\arg\max_{b\in\mathcal{B}}p_{\theta_{\mathrm{beh}}}(b\mid \mathbf{Y},\rho).
\end{equation}
To keep the formulation explicit and reproducible, a matrix-parameterized softmax model can be used, for example
\begin{equation}
\mathbf{s}_{\mathrm{priv}}=\mathbf{W}_{\mathrm{priv}}\mathbf{y}+\mathbf{c}_{\mathrm{priv}},\qquad
p_{\theta_{\mathrm{priv}}}(u\mid \mathbf{y},\rho)=\mathrm{softmax}(\mathbf{s}_{\mathrm{priv}})_u,
\end{equation}
with $\mathbf{W}_{\mathrm{priv}}\in\mathbb{R}^{|\mathcal{U}|\times M}$. Behaviour inference is defined analogously using a sequence feature mapping $\mathbf{f}(\mathbf{Y})\in\mathbb{R}^{d}$ and a weight matrix $\mathbf{W}_{\mathrm{beh}}\in\mathbb{R}^{|\mathcal{B}|\times d}$.

\subsection{Reconstruction as an interpretability check}
Reconstruction is used to visually verify whether coarse body structure and motion cues can be retained while identity-relevant facial details remain unrecoverable at certain sampling rates. We pose reconstruction as a regularized inverse problem
\begin{equation}
\hat{\mathbf{x}}_{\rho}
=
\arg\min_{\mathbf{x}}
\frac{1}{2}\|\mathbf{y}-\mathbf{\Phi}_M\mathbf{x}\|_2^2
+\frac{\lambda}{2}\|\mathbf{L}\mathbf{x}\|_2^2,
\end{equation}
which yields the linear system
\begin{equation}
(\mathbf{\Phi}_M^\top\mathbf{\Phi}_M+\lambda \mathbf{L}^\top\mathbf{L})\hat{\mathbf{x}}_{\rho}
=
\mathbf{\Phi}_M^\top\mathbf{y}.
\end{equation}
The data-consistency gradient is
\begin{equation}
\nabla_{\mathbf{x}}\Big(\tfrac{1}{2}\|\mathbf{y}-\mathbf{\Phi}_M\mathbf{x}\|_2^2\Big)
=\mathbf{\Phi}_M^\top(\mathbf{\Phi}_M\mathbf{x}-\mathbf{y}).
\end{equation}
To emphasize edges and suppress fine textures, we also consider gradient-sparsity regularization
\begin{equation}
\hat{\mathbf{x}}_{\rho}=
\arg\min_{\mathbf{x}}
\frac{1}{2}\|\mathbf{y}-\mathbf{\Phi}_M\mathbf{x}\|_2^2
+\lambda\|\mathbf{D}\mathbf{x}\|_1,
\end{equation}
where $\mathbf{D}$ is a discrete gradient operator matrix.

\subsection{Matrix diagnostics of the sensing operator}
To characterize how sampling rate changes information throughput, we analyze the sensing operator with standard matrix diagnostics. Define the Gram matrices
\begin{equation}
\mathbf{G}_{\Phi}=\mathbf{\Phi}_M\mathbf{\Phi}_M^\top\in\mathbb{R}^{M\times M},\qquad
\mathbf{H}_{\Phi}=\mathbf{\Phi}_M^\top\mathbf{\Phi}_M\in\mathbb{R}^{N\times N},
\end{equation}
and the singular value decomposition
\begin{equation}
\mathbf{\Phi}_M=\mathbf{U}\mathbf{\Sigma}\mathbf{V}^\top,\qquad
\mathbf{\Sigma}=\mathrm{diag}(\sigma_1,\dots,\sigma_M).
\end{equation}
As $\rho$ increases, the effective rank and the spectral mass of $\mathbf{\Sigma}$ typically increase, improving stability of inversion and increasing discriminative content available in measurement-domain inference.

\subsection{Recognisability, critical sampling rates, and existence of a safe interval}
The purpose here is not to pinpoint critical sampling rates to a single numerical value, but to establish the existence of two thresholds whose ordering creates a nonempty interval in which behaviour remains recognisable while privacy remains unrecognisable.

For sampling rate $\rho$, define the Bayes-optimal correct classification probabilities
\begin{equation}
A_{\mathrm{priv}}(\rho)\triangleq \sup_{g}\ \mathbb{P}\big(g(\mathbf{y})=U\mid \rho\big),\qquad
A_{\mathrm{beh}}(\rho)\triangleq \sup_{h}\ \mathbb{P}\big(h(\mathbf{Y})=B\mid \rho\big),
\end{equation}
where the suprema are over all measurable decision functions. Define the privacy advantage over chance level
\begin{equation}
\mathrm{Adv}_{\mathrm{priv}}(\rho)\triangleq A_{\mathrm{priv}}(\rho)-\frac{1}{|\mathcal{U}|}.
\end{equation}
For very small $\rho$, the measurement vector contains only a few projections and the posterior over identities approaches the prior under zero-mean sensing and noise, so $\mathrm{Adv}_{\mathrm{priv}}(\rho)$ can be made arbitrarily close to zero. For sufficiently large $\rho$, both tasks may achieve high correct rates because the sensing operator preserves more scene information. The nesting of measurements as $M$ increases implies that both $A_{\mathrm{priv}}(\rho)$ and $A_{\mathrm{beh}}(\rho)$ are nondecreasing in $\rho$.

Given thresholds $0<\beta_{\mathrm{priv}}<1$ and $0<\alpha_{\mathrm{beh}}<1$, define critical sampling rates by
\begin{equation}
\rho_{\mathrm{priv}}^\star=\sup\{\rho:\ A_{\mathrm{priv}}(\rho)\le \beta_{\mathrm{priv}}\},\qquad
\rho_{\mathrm{beh}}^\star=\inf\{\rho:\ A_{\mathrm{beh}}(\rho)\ge \alpha_{\mathrm{beh}}\}.
\end{equation}
If
\begin{equation}
\rho_{\mathrm{beh}}^\star<\rho_{\mathrm{priv}}^\star,
\end{equation}
then the safe interval
\begin{equation}
\mathcal{I}_{\mathrm{safe}}=[\rho_{\mathrm{beh}}^\star,\ \rho_{\mathrm{priv}}^\star]
\end{equation}
is nonempty. For any $\rho\in\mathcal{I}_{\mathrm{safe}}$, the two inequalities hold simultaneously:
\begin{equation}
A_{\mathrm{beh}}(\rho)\ge \alpha_{\mathrm{beh}},\qquad
A_{\mathrm{priv}}(\rho)\le \beta_{\mathrm{priv}}.
\end{equation}
The experimental sampling rates are chosen to lie within such an interval, so that risk-related behaviour can be detected while face identity cannot be reliably inferred.

\subsection{Structural rationale for the ordering of thresholds}
We formalize why behaviour can become recognisable at lower sampling rates than face identity. Let $\mathcal{S}_{\mathrm{beh}}\subset\mathbb{R}^{N}$ denote a subspace capturing variations relevant to pose and action, and let $\mathcal{S}_{\mathrm{priv}}\subset\mathbb{R}^{N}$ denote a higher-complexity subspace capturing identity-relevant facial variations. Assume
\begin{equation}
d_{\mathrm{beh}}=\dim(\mathcal{S}_{\mathrm{beh}})\ll d_{\mathrm{priv}}=\dim(\mathcal{S}_{\mathrm{priv}}).
\end{equation}
If the sensing operator provides an approximately isometric embedding on a subspace $\mathcal{S}$, meaning there exist constants $0<c_1<c_2$ such that
\begin{equation}
c_1\|\mathbf{z}\|_2^2\le \|\mathbf{\Phi}_M\mathbf{z}\|_2^2\le c_2\|\mathbf{z}\|_2^2,\qquad \forall \mathbf{z}\in\mathcal{S},
\end{equation}
then the corresponding structure is preserved in the measurement domain and supports stable discrimination. Achieving such stability typically requires a number of measurements that scales with the subspace dimension, hence fewer measurements suffice for $\mathcal{S}_{\mathrm{beh}}$ than for $\mathcal{S}_{\mathrm{priv}}$. This yields $\rho_{\mathrm{beh}}^\star<\rho_{\mathrm{priv}}^\star$ and supports the existence of a nonempty interval where behaviour is recognisable while privacy remains protected.

\subsection{Reproducibility}
Across sampling rates, we keep the pattern library, optical geometry, illumination power, integration time, calibration procedure, and preprocessing operators fixed. Only $M$ is changed, so the effect of $\rho$ is isolated to the sensing operator dimension through $\mathbf{\Phi}_M=\mathbf{S}\mathbf{\Phi}$. The same statistical definitions of recognisability and the same decision rules are applied across all sampling rates. The deployment setting selects a sampling rate within a safe interval $\mathcal{I}_{\mathrm{safe}}$, which does not require numerically pinpointing either threshold, but requires that the interval is nonempty and that the operating point lies inside it.

\appendix

\section{Synthetic Privacy-Sensitive Behavior Dataset}
\label{syn_dataset}

\subsection{Scene and Activity Taxonomy}

\begin{table}[h]
\centering
\caption{\textbf{Privacy-sensitive scenes and activity categories in the synthetic dataset.}
Each scene includes normal and abnormal behaviors that are difficult or unethical to capture using conventional surveillance.}
\label{tab:synthetic_dataset_scenes}
\setlength{\tabcolsep}{8pt}
\renewcommand{\arraystretch}{1.15}
\begin{tabularx}{\textwidth}{l X X}
\toprule
\textbf{Scene} & \textbf{Normal Activities} & \textbf{Abnormal Activities} \\
\midrule
Public restroom &
Using a toilet stall; washing hands; tidying clothes; applying makeup; standing near the entrance &
Facility vandalism; collapse due to illness; drug-related behaviors; covert photographing; physical fighting \\
\midrule
Changing room /
fitting room &
Changing clothes; trying on garments; looking in a mirror; brief conversation &
Shoplifting; facility damage; harassing postures; physical fighting \\
\midrule
Public shower /
bathhouse &
Showering; scrubbing oneself; standing under the shower; brief conversation &
Covert photographing; harassing postures; facility damage; collapse due to illness; physical fighting \\
\bottomrule
\end{tabularx}
\end{table}

\subsection{Prompt Template for Synthetic Image Generation}

\begin{table}[h]
\centering
\caption{\textbf{Prompt template used for generating privacy-sensitive synthetic images.}
The template enforces identity anonymization while preserving coarse behavioral and contextual cues.}
\label{tab:prompt_template}
\setlength{\tabcolsep}{10pt}
\renewcommand{\arraystretch}{1.15}
\begin{tabularx}{\textwidth}{X}
\toprule
\textbf{Prompt Template} \\
\midrule
High-definition overhead surveillance camera view, focused on \{scene\_description\}. 
There are \{person\_count\_text\} in the scene, and they are performing \{activity\_description\}. 
The generated image should only maintain the basic facial features of humans, with anonymized dummy faces to avoid privacy issues. 
Sensitive body parts can be blurred. \\
\bottomrule
\end{tabularx}
\end{table}

\bibliographystyle{plainnat}
\bibliography{paper}

\begin{thebibliography}{12}
\providecommand{\natexlab}[1]{#1}
\providecommand{\url}[1]{\texttt{#1}}
\expandafter\ifx\csname urlstyle\endcsname\relax
  \providecommand{\doi}[1]{doi: #1}\else
  \providecommand{\doi}{doi: \begingroup \urlstyle{rm}\Url}\fi

\bibitem[An et~al.(2025)An, Hu, Huang, Huang, Li, Liang, Shao, Song, Wang, Yuan, et~al.]{an2025ai}
Hongjun An, Wenhan Hu, Sida Huang, Siqi Huang, Ruanjun Li, Yuanzhi Liang, Jiawei Shao, Yiliang Song, Zihan Wang, Cheng Yuan, et~al.
\newblock Ai flow: Perspectives, scenarios, and approaches.
\newblock \emph{arXiv preprint arXiv:2506.12479}, 2025.

\bibitem[Bai et~al.(2025)Bai, Cai, Chen, Chen, Chen, Cheng, Deng, Ding, Gao, Ge, Ge, Guo, Huang, Huang, Huang, Hui, Jiang, Li, Li, Li, Li, Lin, Lin, Liu, Liu, Liu, Liu, Liu, Liu, Lu, Luo, Lv, Men, Meng, Ren, Ren, Song, Sun, Tang, Tu, Wan, Wang, Wang, Wang, Wang, Xie, Xu, Xu, Xu, Yang, Yang, Yang, Yang, Yu, Zhang, Zhang, Zhang, Zheng, Zhong, Zhou, Zhou, Zhou, Zhu, and Zhu]{Qwen3-VL}
Shuai Bai, Yuxuan Cai, Ruizhe Chen, Keqin Chen, Xionghui Chen, Zesen Cheng, Lianghao Deng, Wei Ding, Chang Gao, Chunjiang Ge, Wenbin Ge, Zhifang Guo, Qidong Huang, Jie Huang, Fei Huang, Binyuan Hui, Shutong Jiang, Zhaohai Li, Mingsheng Li, Mei Li, Kaixin Li, Zicheng Lin, Junyang Lin, Xuejing Liu, Jiawei Liu, Chenglong Liu, Yang Liu, Dayiheng Liu, Shixuan Liu, Dunjie Lu, Ruilin Luo, Chenxu Lv, Rui Men, Lingchen Meng, Xuancheng Ren, Xingzhang Ren, Sibo Song, Yuchong Sun, Jun Tang, Jianhong Tu, Jianqiang Wan, Peng Wang, Pengfei Wang, Qiuyue Wang, Yuxuan Wang, Tianbao Xie, Yiheng Xu, Haiyang Xu, Jin Xu, Zhibo Yang, Mingkun Yang, Jianxin Yang, An~Yang, Bowen Yu, Fei Zhang, Hang Zhang, Xi~Zhang, Bo~Zheng, Humen Zhong, Jingren Zhou, Fan Zhou, Jing Zhou, Yuanzhi Zhu, and Ke~Zhu.
\newblock Qwen3-vl technical report.
\newblock \emph{arXiv preprint arXiv:2511.21631}, 2025.

\bibitem[Chen et~al.(2023)Chen, Sun, Li, and Li]{chen2023computational}
Yifan Chen, Zhe Sun, Chen Li, and Xuelong Li.
\newblock Computational ghost imaging in turbulent water based on self-supervised information extraction network.
\newblock \emph{Optics \& Laser Technology}, 167:\penalty0 109735, 2023.

\bibitem[Chen et~al.(2025)Chen, An, Sun, Tian, Chen, Spielmann, and Li]{chen2025large}
Yifan Chen, Hongjun An, Zhe Sun, Tong Tian, Mingliang Chen, Christian Spielmann, and Xuelong Li.
\newblock Large model enhanced computational ghost imaging.
\newblock \emph{Science China Technological Sciences}, 68\penalty0 (11):\penalty0 2120403, 2025.

\bibitem[Duarte et~al.(2008)Duarte, Davenport, Takhar, Laska, Sun, Kelly, and Baraniuk]{duarte2008single}
Marco~F Duarte, Mark~A Davenport, Dharmpal Takhar, Jason~N Laska, Ting Sun, Kevin~F Kelly, and Richard~G Baraniuk.
\newblock Single-pixel imaging via compressive sampling.
\newblock \emph{IEEE Signal Processing Magazine}, 25\penalty0 (2):\penalty0 83--91, 2008.

\bibitem[{Google DeepMind}(2023)]{TeamGemini2023}
{Google DeepMind}.
\newblock Gemini: A family of highly capable multimodal models.
\newblock \emph{arXiv preprint arXiv:2312.11805}, 2023.

\bibitem[Huang et~al.(2007)Huang, Ramesh, Berg, and Learned-Miller]{Huang2007LFW}
Gary~B. Huang, Manu Ramesh, Tamara Berg, and Erik Learned-Miller.
\newblock Labeled faces in the wild: A database for studying face recognition in unconstrained environments.
\newblock In \emph{Proceedings of the IEEE International Conference on Automatic Face and Gesture Recognition (FG)}, pages 236--243, 2007.

\bibitem[Li et~al.(2024)Li, Chen, Tian, and Sun]{li2024part}
Xuelong Li, Yifan Chen, Tong Tian, and Zhe Sun.
\newblock Part-based image-loop network for single-pixel imaging.
\newblock \emph{Optics \& Laser Technology}, 168:\penalty0 109917, 2024.

\bibitem[Lin(2004)]{lin-2004-rouge}
Chin-Yew Lin.
\newblock {ROUGE}: A package for automatic evaluation of summaries.
\newblock In \emph{Text Summarization Branches Out}, pages 74--81, Barcelona, Spain, July 2004. Association for Computational Linguistics.
\newblock URL \url{https://aclanthology.org/W04-1013/}.

\bibitem[Lu and Tang(2014)]{LuTang2014GaussianFace}
Chaochao Lu and Xiaoou Tang.
\newblock Surpassing human-level face verification performance on lfw with gaussianface.
\newblock arXiv preprint, 2014.
\newblock URL \url{https://arxiv.org/abs/1404.3840}.

\bibitem[{OpenAI}(2025)]{OpenAI2025GPT5}
{OpenAI}.
\newblock Gpt-5 system card.
\newblock Technical report, 2025.
\newblock URL \url{https://openai.com/research/gpt-5-system-card}.
\newblock Accessed 2026-01-20.

\bibitem[{World Health Organization}(2024)]{WHO2024YouthViolence}
{World Health Organization}.
\newblock Youth violence.
\newblock WHO Newsroom Fact sheet, October 2024.
\newblock URL \url{https://www.who.int/news-room/fact-sheets/detail/youth-violence}.
\newblock Accessed 2026-01-20.

\end{thebibliography}

\end{document}